\documentclass[manuscript]{acmart}
\usepackage{subfig}
\usepackage{hyperref}

\AtBeginDocument{%
  \providecommand\BibTeX{{%
    \normalfont B\kern-0.5em{\scshape i\kern-0.25em b}\kern-0.8em\TeX}}}

\copyrightyear{2022}
\acmYear{2022}
\setcopyright{rightsretained}
\acmConference[FAccT '22]{2022 ACM Conference on Fairness, Accountability, and Transparency}{June 21--24, 2022}{Seoul, Republic of Korea}
\acmBooktitle{2022 ACM Conference on Fairness, Accountability, and Transparency (FAccT '22), June 21--24, 2022, Seoul, Republic of Korea}\acmDOI{10.1145/3531146.3533095}
\acmISBN{978-1-4503-9352-2/22/06}

\begin{document}

\title{Social Inclusion in Curated Contexts: Insights from Museum Practices}


\author{Han-Yin Huang}
\email{H.Huang-4@tudelft.nl}
\affiliation{
\institution{Delft University of Technology}
\country{The Netherlands}
}
\author{Cynthia C.~S. Liem}
\email{C.C.S.Liem@tudelft.nl}
\affiliation{
\institution{Delft University of Technology}
\country{The Netherlands}
}


\begin{abstract}

Artificial intelligence literature suggests that minority and fragile communities in society can be negatively impacted by machine learning algorithms due to inherent biases in the design process, which lead to socially exclusive decisions and policies. Faced with similar challenges in dealing with an increasingly diversified audience, the museum sector has seen changes in theory and practice, particularly in the areas of representation and meaning-making. While rarity and grandeur used to be at the centre stage of the early museum practices, folk life and museums' relationships with the diverse communities they serve become a widely integrated part of the contemporary practices. These changes address issues of diversity and accessibility in order to offer more socially inclusive services. Drawing on these changes and reflecting back on the AI world, we argue that the museum experience provides useful lessons for building AI with socially inclusive approaches, especially in situations in which both a collection and access to it will need to be curated or filtered, as frequently happens in search engines, recommender systems and digital libraries. We highlight three principles: (1) Instead of upholding the value of neutrality, practitioners are aware of the influences of their own backgrounds and those of others on their work. By not claiming to be neutral but practising cultural humility, the chances of addressing potential biases can be increased. (2) There should be room for situational interpretation beyond the stages of data collection and machine learning. Before applying models and predictions, the contexts in which relevant parties exist should be taken into account.  (3) Community participation serves the needs of communities and has the added benefit of bringing practitioners and communities together.
\end{abstract}



\maketitle


\section{Introduction}\label{introduction}
In many applications of AI, machine learning technology is employed to refocus the attention of humans. In many classification tasks, the intention is to automatically label digital objects in ways that scale beyond human capacity. The necessary standardisation in this procedure has both been lauded for its efficiency, yet criticised for its taxonomic crudeness, which tends to be particularly oblivious to cultural and contextual sensitivities and minority perspectives.

Further downstream, machine learning technology empowers how vast collections of automatically describable digital objects get organised and re-organised, and how, when a human seeks to interact with the collections, they will be filtered down to much smaller-sized `relevant' sub-selections. In digital libraries, search engines and recommender systems alike, this has again both been lauded and criticised. On the positive side, automated filtering technologies can traverse digital collections more broadly and efficiently than humans can do. Generally, through digitalisation and unified access mechanisms, collection items that would be hard to reach in physical archives can now universally and globally be retrieved. Yet, on the negative side, filtering and access mechanisms in our digital information space may not actually capitalise on the diversity they could in principle offer. Many services do not naturally stimulate their audiences to venture beyond the filtered sub-selections offered to them, and perspectives provoking the strongest engagement easily get prioritised, at the risk of favoring `majority-group' or heavily polarising world views.

Often, we face large collections reflecting a pluriformity of perspectives and interpretations, while people interested in interacting with these collections will only get to see limited selections of items. How can such selections be constituted and presented in ways that do right to the original diversity in the broader collections? How can curation be performed in ways that respect, engage and include audiences beyond mainstream perspectives, without necessarily going down populist routes? In digital information access applications, we are starting to recognise these are relevant, yet still unsolved challenges. When not applied consciously, AI technology may however actively work against these challenges, and it still is an open question how to avoid this. However, in the physical world, over the last decades, similar questions and challenges have actually been posed and addressed in the museum community.


For at least two decades, the museum sector has been confronted with debates on its social role, potential biases in its work and issues of social inclusion and cultural diversity~\cite{matarasso,NewmanMcLean1998, DoddandSandell2001}. According to the current museum definition of the International Council for Museums (ICOM), a museum is an institution that `acquires, conserves, researches, communicates and exhibits' both the intangible and tangible heritage~\cite{ICOMdefinition}. In this process, a museum decides what is accepted into its collection and what is deemed significant heritage for the past, present and future generations. Museums are, therefore, institutions with the power to endorse views.

Traditionally, museums are seen as places that house and exhibit items of historic, scientific or aesthetic values, where the criteria primarily have been based on the dominant groups whose cultures museums were founded on. Such a curating process can actually be interpreted as one of exclusion (i.e., deciding what does \emph{not} get to be accepted into a collection, and what thus would not be deemed significant heritage). Recognising this, various activities and actions have been initiated to still foster social inclusion, as will be discussed in this paper.

Beyond these various initiatives in different museums, a recent debate surrounding the museum definition proposal of ICOM illustrates the sector's attempt to officially formalise inclusive practices as part of its raison d'\^{e}tre. Before its general assembly in 2019, ICOM announced that it would put a new museum definition proposal to be voted on during the assembly. This proposed new definition referred to museums as `democratising, inclusive and polyphonic spaces', whose aims are to `contribute to human dignity and social justice, global equality and planetary wellbeing'~\cite{ICOMannouncement}. While applauded by supporters, the announcement was also met with criticisms, among which the lack of transparency in the decision-making process and insufficient consultation with members and national committees came to the fore. The 2019 ICOM general assembly closed with a postponement of the vote. A series of consultation sessions, led by the ICOM Define committee and scheduled to take place until the first quarter of 2022, ensues. According to the analysis and report of the ICOM Define Committee's second consultation with national committees, among the terms mentioned by national committees, many are related to social inclusion and diversity, for example, `inclusive' is mentioned by 66 percent of the national committees, `open to society/public' 52 percent, `community/society' 51 percent, `accessibility' 45 percent, `service to society' 44 percent and `diversity' 41 percent~\cite{ICOMDefine2021}. Among the five draft proposals for museum definitions put forward by ICOM in February 2022, museums being inclusive and accessible to a variety of audiences is mentioned in all five definition proposals\footnote{\url{https://icom.museum/en/news/on-the-way-to-a-new-museum-definition-we-are-doing-it-together/}, retrieved April 30, 2022. National committees have been asked to identify their preferred proposals in the fourth and final stage of consultation. The final report is scheduled to be published in May 2022.}.

Both in the museum sector and in AI-powered digital information technology, potential biases in curated contexts pose very tangible challenges. Both worlds have a similar process, that starts with collecting what is deemed relevant to their purposes: in the museum's case, the tangible and intangible heritage, and in case of AI technology, the data. The collections are then interpreted, after which selected items and views are presented to audiences. Drawing on this parallel, we argue that the existing debates and reform movements with regard to the museum experience can provide useful insights for developing socially inclusive approaches in AI.

In their work on Lessons from Archives, Jo and Gebru draw attention to how sociocultural data is collected and processed in archives~\cite{JoGebru2020archives}. They identify principles that the machine learning community can learn from, in particular on practices to increase transparency with regard to data collection.
In this paper, we complement these insights by taking a broader process and organisational perspective, looking beyond the collection phase, and also including the interpretation, application and presentation phases, that ultimately reach human audiences.


\section{Social Inclusion in Museums}\label{SocialInclusioninMuseums}
Although, because of its more positive note, \emph{social inclusion} is a term applied more often in policy and in practice, it is also used interchangeably with social exclusion, social cohesion, solidarity, integration and social capital~\cite{NewmanMcClean2004presumptionl,Silver2015}. First coined in France in the 1970s, \emph{social exclusion} referred to the marginalised groups of the French welfare system and has since then become part of the European political discussions. As time went by, the discussion in policy moved to the more positive `social inclusion'~\cite{Silver2015}. According to Silver, social inclusion has different connotations and applications in different settings on the national and neighbourhood levels, and is also located in place~\cite{Silver2015}. Social inclusion is therefore a context-specific term. To many cultural institutions, what social inclusion entails is extremely fluid, and it thus is difficult to give a definition that fits all practices~\cite{NewmanMcClean2004presumptionl,Silver2015}.

Richard Sandell is one of the early museologists who champions museums as `agents of social inclusion’ and suggests that museums nowadays are involved in or committed to different degrees of socially inclusive practices and ideals~\cite{sandell1998agents}. The inclusive museums engage from the cultural dimension by tackling representation and access aspects in museum practice. Museums as agents of social regeneration take on the social, economic and political dimensions and seek to promote better quality of lives for individuals, also encouraging personal development of the disadvantaged groups. Finally, museums as vehicles for broad social change aim to encourage positive social change and open themselves up as forums for public debates~\cite{sandell1998agents}. Social inclusion in museum practice responds at first to the discussions of museum accessibility for a wider range of visitors (cultural dimension) and later includes the discussion the capacity of museums empowering its visitors (social, economic and political dimensions).

Engaging social inclusion in the museum sector has become an important part of the discussion about democratising the museum practice~\cite{sandell2003sectoral}. In contrast to its traditional image being situated in a power structure that posits museum staff as experts and a top-down practice model, the democratised museum practice seeks to revisit accessibility to the museum not just in terms of physical access, but also in terms of social, cultural and economic access, changing its relationship with the communities that have long been underrepresented by the museum sector. Democratised museum thinking is manifest in the changes in the theorising of representation and meaning-making in museum practices. As Mason outlines in the representational discourse of social inclusion in museums practices, there are three strands of influences that intertwine to make socially inclusive practice an important aspect in the contemporary museum practice~\cite{mason2004discourse}. As the targeted audience shifts, from the elite educated and knowledgeable group to a more general public without specialist knowledge, subject matters also shift to include folk life and everyday collecting: the ordinary heritage that the visitors can recognise themselves in. A second strand is the shift of Culture with capital C to culture with small letter c, from the elitist culture to simply referring a way of life. The third strand is the influence of the environmental movements in the 1960s and discussions on these in the museum sector, in the form of new museology~\cite{vergo1989} for \emph{ecomuseology}~\cite{davis2011}. At the core of the ecomuseum philosophy is the museum's thinking beyond the confinement of its four walls, building relationships with communities.

Development in museum learning theories also informs new perspectives in the meaning-making of museum visits. Literature of how museum visitors make sense of the messages conveyed in exhibitions and programmes suggests that there is no one universal manner of museum visit experience. Visitors go to museums for different purposes with various expectations~\cite{FalkDierking2013, HooperGreenhill2000}. It is therefore impossible for museums to create a `one thing fits all' programme.  The visitors who come through the gate also bring with them their own personal, cultural, social and political context. Every visitor can experience a museum visit very differently.

To illustrate how these notions were implemented for socially inclusive museum experiences, we discuss three key principles, which in our perspective can be translated to parallels in AI practice. First of all, we describe how instead of setting neutrality as the ultimate goal, an attitudinal change informed by cultural humility may provide better guidance. Secondly, allowing room for situational interpretation at different stages of the development process ensures that different contexts are considered and acknowledged. This implies that potential biases should keep being readdressed throughout the whole chain of services; that is, not only at the data collection, but also during the design, interpretation and implementation stages. Lastly, it is important to actively engage communities so that the practice connects with the community experience.

\section{The Museum Insights}\label{sec:MuseumInsights}
\subsection{Neutrality Revisited}\label{neutralitymuseum}
As discussed earlier, to become more socially inclusive, a sectoral change is needed to transform the traditional museum practice. This does not trivially happen. Among the aspects identified by Sandell that inhibit the sectoral change, he first points out the attitudes of museum workers that do not subscribe to social inclusion~\cite{sandell2003sectoral}. To facilitate attitude changes, Sandell suggests training programmes based on first-hand experience and guided by the groups and communities, that have traditionally not seen themselves represented in the museums. Such attitude change can be summarised as `cultural humility'. The concept of \emph{cultural humility} has originally been promoted in health care and library and information science, to tackle oppressive descriptions in the medical and archival practices, respectively. It refers to the ability to acknowledge the influences personal backgrounds could have on the practitioners and their practices, and finding appropriate approaches to deal with these influences~\cite{tai2020}.

Museums do not exist in a vacuum and neither do the visitors. As Stijn Schoonderwoerd, the general director of National Museum of World Cultures (NMVW),  points out, ``Museums, although they are often seen as places solely dedicated to beautiful things, and therefore as neutral and non-political spaces, are players in the social and political arena too''~\cite{Schoonderwoerd2020}. A museum collects, interprets and displays tangible and intangible heritage in its care. Its decision-making process illustrates a perspective curated and presented by the museum. The key word here is \emph{a} perspective, rather than \emph{the} perspective. In this process, a museum makes decisions about what the museum deems relevant to the institution in its exhibitions and education and outreach programmes. Museums function, therefore, in curated contexts that are not without implications of certain values.

Instead of emphasising how `neutral' museums should be, an attitude of `cultural humility' signifies changes in the curatorial practices of museums. To facilitate an inclusive curatorial practice, the selection processes to decide what goes in the collections are informed by a museum’s mission statement and collection policy. The collection policy will be reviewed at a regular interval to make sure the museum does not collect too many repeated, similar items, and the policy keeps up with developments within the museum. Changes can also be seen in the calls for decolonising museum narratives, that go from seeing different cultures as the exotic others, to placing different cultures in their own contexts. History and the heritage of the ruling and dominant groups are not the only heritage considered `qualified' for museum collections.

One of the most telling examples in the museums' efforts of putting cultural humility to practice, is the re-examination of terminology used in the museum sector. When an item is accepted into the collection,  museum curators conduct research into its material culture such as its origin, material, artist, maker and previous collectors, and update relevant findings and literature related to the item. Derogatory terms in the catalogue descriptions and exhibition texts that are outdated in contemporary society are being identified and updated through research projects in museum terminology. A good example of this is the Terminology project by the Amsterdam Rijksmuseum\footnote{\url{https://www.rijksmuseum.nl/en/research/our-research/overarching/terminology}, retrieved April 30, 2022.}.

\begin{figure}
\includegraphics[width=0.7\columnwidth]{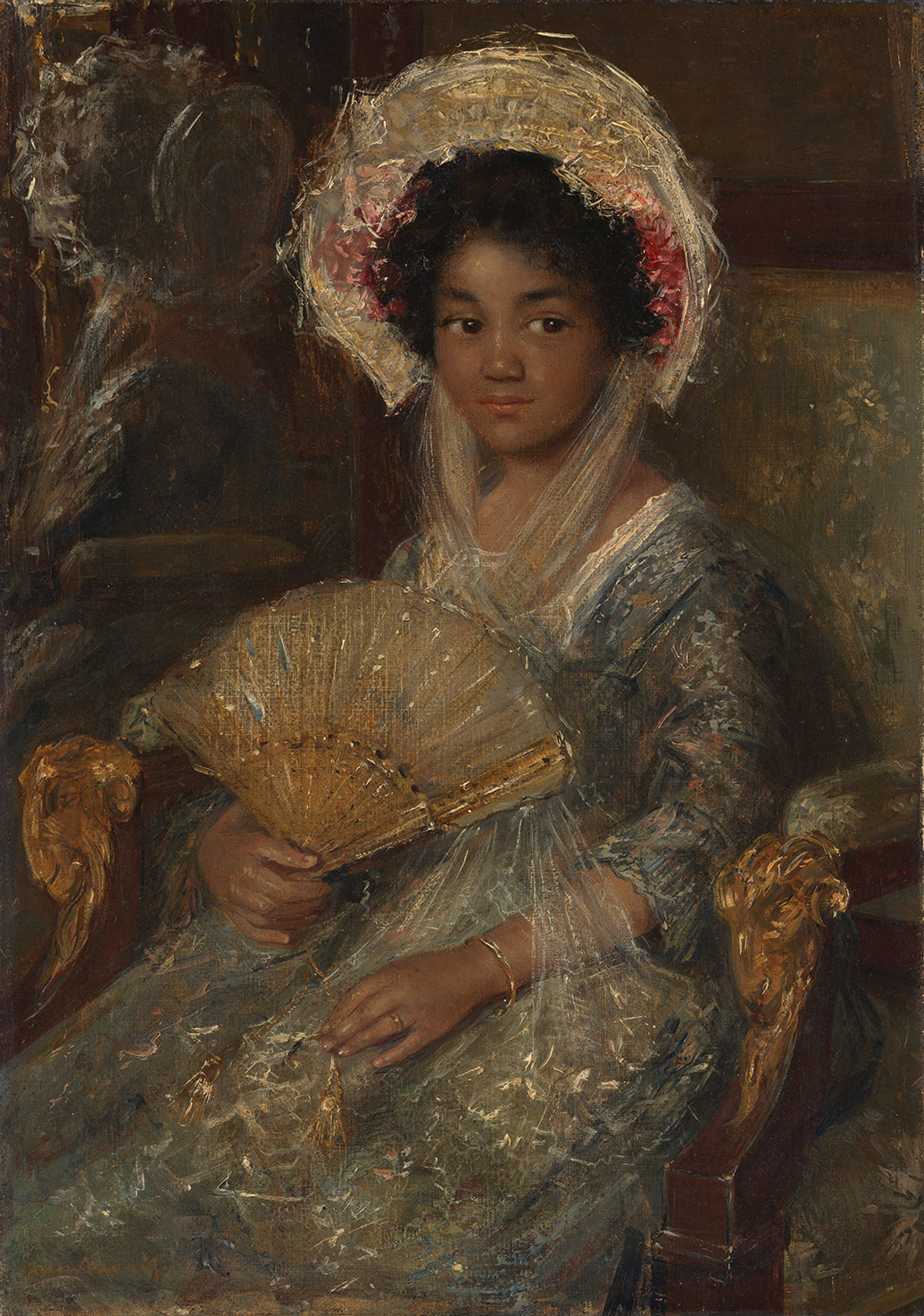}
\caption{\emph{Isabella}, Simon Maris, c.\ 1906, oil on canvas. Rijksmuseum, Amsterdam.}\label{fig_isabella}
\end{figure}

The Rijksmuseum's Terminology project started in 2015 with the painting shown in Figure~\ref{fig_isabella}. When the painting became part of the museum's collection in 1922, it was described as ``East Indian type. Oriental girl sitting in an armchair''. In the 1970s, the work was titled ``Little Negress'' by one of the curators then. In 2015, this title set off a public debate; as a result, the Museum assembled an inter-departmental team tasked to investigate the language used by the museum. Although the task force originally gave the painting a new title ``Young Woman with a Fan'', further research into the documents and communications in the Maris family archives unveiled in 2020 that the girl's first name is Isabella, and that she was 12 years old when she posed for Maris. ``Isabella'' sparked a series of term changes in the museum, for instance, `slaves' are now described as `enslaved' and `koelies' are now referred to as `contract workers'. Eurocentric terms have also been identified; for example, `Indians' are now referred to as 'original residents of America' and `exotic', an adjective used to describe someone different from the European norm, is now avoided by the museum.

 Similarly, it is worthwhile to highlight the Words Matter project by the Dutch National Museum of World Cultures\footnote{\url{https://www.materialculture.nl/en/publications/words-matter}, retrieved April 30, 2022.}. This project produced a work-in-progress document, compiling a list of terms that were previously used but are now identified as derogatory, accompanied by various commenting essays~\cite{WordsMatter}. It also discusses how terminology used in museums have impacts on its work; not just in collection, but also in education and interpretation, and its relationship to the community. Wayne Modest, one of the authors in the Words Matter document, says in response to controversies in the Netherlands about changing words and texts, that ``Paying attention to words means acknowledging that the words we use affect a person or a group feels excluded or included, whether they feel a sense of belonging to society. This is about representation, recognition and respect.''~\cite{WayneModest2020}

\subsection{Situational Interpretation}\label{contextmuseum}
As mentioned in Section~\ref{SocialInclusioninMuseums}, social inclusion is a fluid concept and its definition and practice differ according to the context in which it is applied. The museum practice of social inclusion goes beyond ensuring that a collection has a diverse representation of its communities and audiences: beyond the collection and curatorial departments, other areas of the museum practice, such as the education and outreach departments, actively engage in offering inclusive services. Different parts of the museum services provide situational interpretation to respond to the needs of the audience groups they are catering for.

Apart from changing the outdated languages used in the museum sector, as outlined in Section~\ref{neutralitymuseum}, changes can be seen in how curators produce interpretive texts of museum exhibitions with more visitor-oriented approaches. Accessible texts that are easy for the public to read and relate to, instead of jargon-loaded academic texts, are considered more effective in getting the messages across. Moreover, there is no guarantee that the texts and messages will be perceived and understood to the same degree and effect by all visitors~\cite{HooperGreenhill2000}. Understanding that meanings of museum displays can be shifting from one visitor to another, museums start exploring the inclusion of texts written by community members, and the engagement of visitors in reflective questions in exhibitions. In New York, for example, an exhibition by the New York Historical Society, Scenes of New York City, displays art works alongside texts written by New Yorkers who are connected to the subject matters of the works on display\footnote{\url{https://www.nyhistory.org/exhibitions/scenes-of-new-york-city-elie-and-sarah-hirschfeld}, retrieved April 30, 2022.}. For an oil painting by Gifford Beal which features a top-hat-wearing horse carriage driver, visitors see both interpretive texts from the exhibition curator, and from a horse carriage driver in Central Park. While the curator's text introduces the painter and touches on the carriage tour history at Central Park, the carriage driver's text is a personal reflection of spring time and being a carriage driver in warm weather in the park, memories triggered by the painting~\cite{JacobsNYTimes2021}.

Similarly, when museum educators plan to utilise an item in collection as part of their education programme or campaign, the research results (such a descriptive texts in the catalogue or labels from past exhibitions) will form the basis of this endeavour. They are, however, often adapted in order to become suitable to the learning needs and specific contexts of the target audience. In the process of developing these resources, the collaboration between the curatorial department and the education department makes sure that the produced texts or knowledge promoted is in line with the research result of the curatorial department, but also incorporates the understandings of the education department on how best to convey messages across to the target group to achieve the desired learning outcomes.

With the museum collection as backbones, situational interpretation enables different parts of the museum services to be bridges between the collection and the visitors, by translating into languages or formats that visitors feel comfortable with. An inclusive museum is not only constituted by an inclusive collection, but includes a whole range of museum services that reach the museum visitors.

\subsection{Community Participation}\label{communitymuseum}
In the effort to offer inclusive practices, the community plays an important role, as informed by the new museology and ecomuseum philosophy. New museology places museums in relationships with the communities in which they exist, and serve at the centre of its discussion. The ecomuseum philosophy expands on this and argues that museums should not be confined to its four walls. Instead of a museum building, an ecomuseum operates within a territory, instead of a building, and has a close tie with the communities whose lives are attached to this territory~\cite{Corsane2007eval}. Museums serve, therefore, as a `contact zone'  where the different contexts, narratives and complexities converge~\cite{Witcomb2003}.

In order to have more diverse community narratives in the museums, audience development becomes an important aspect of museums' public engagement. The lack of a universal definition of what constitutes a diverse and inclusive museum is evident in the audience development practice. For example, the concept of community curators both refers to a method of collecting, and a strategy of audience development and outreach. This way of collecting is usually a project-based, community-oriented approach to collecting. The contexts of this collecting process are documented through recordings of interviews of relevant parties and descriptions, so that these unique decision-making processes are not lost. Such an approach is also a way to build relationships with underrepresented groups.

Collecting Birmingham, for example, was a three-year (2015--2018) project by the Birmingham Museums Trust in the city of Birmingham, UK\footnote{\url{https://dams.birminghammuseums.org.uk/asset-bank/action/browseItems?categoryId=1978&categoryTypeId=2}, retrieved April 30, 2022.}. The first year of the project mainly consisted of consultation sessions with focus groups to discuss different themes surrounding the participants' life experiences in Birmingham. The Trust also recruited Collection Ambassadors who promoted the project in different areas of the city. After the first year, however, the Trust felt that most participants were already people involved in heritage in some way, and looked to expand its participant profiles. In the second year, the project team went to different venues and events in the local neighbourhoods. More contentious objects related to race or religion emerged as a result of the diversity of respondents in the second year~\cite{ContemporaryCollecting}.

\begin{figure}
\includegraphics[width=0.7\columnwidth]{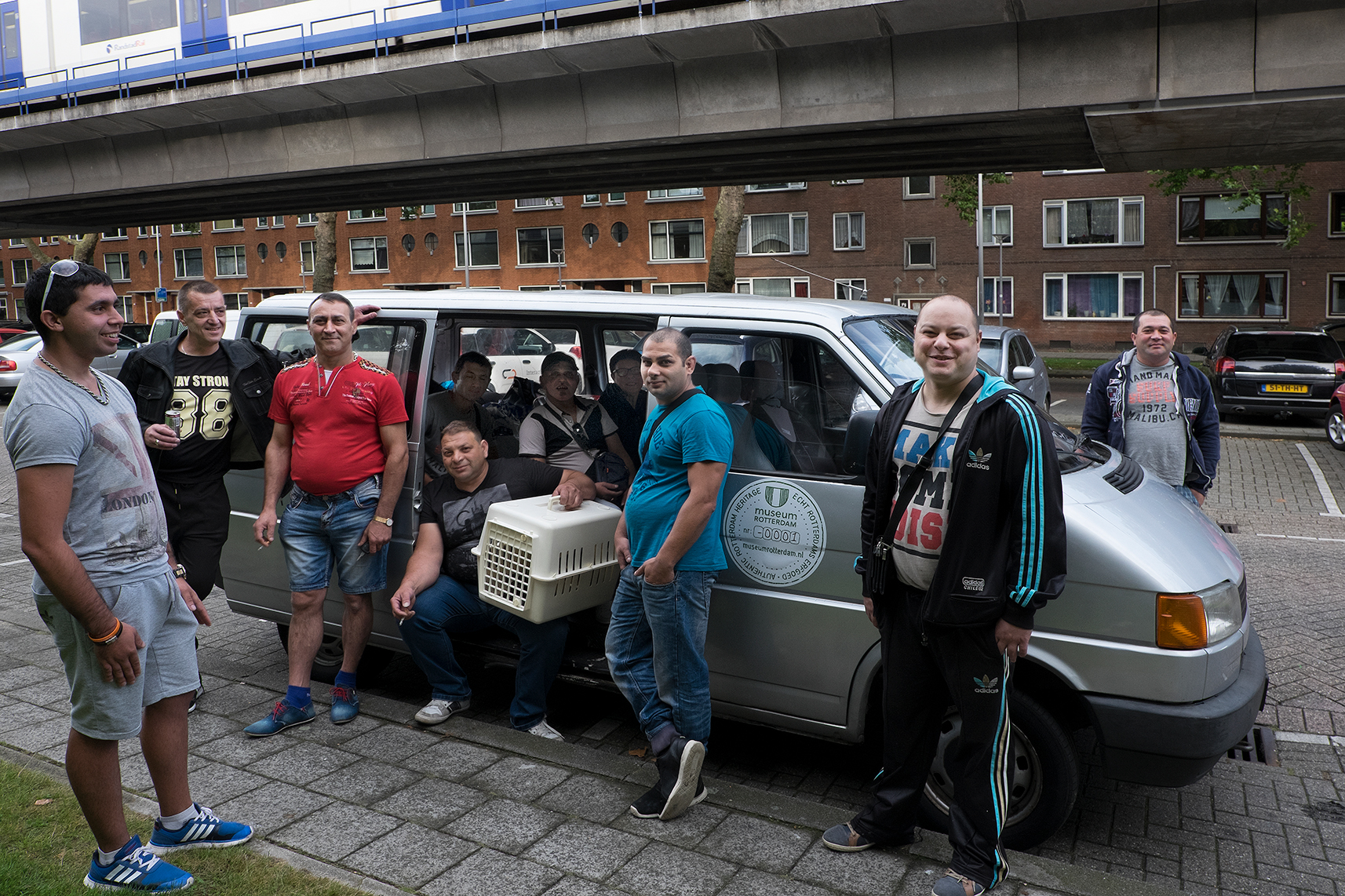}
\caption{\emph{Kamen's minivan}, Object No.\ 0001 of Echt Rotterdams Erfgoed collection. Photo Credit: Joop Reijngoud.}\label{fig_minivan}
\end{figure}

Echt Rotterdams Erfgoed (Authentic Rotterdam Heritage, ERE hereon) is an ongoing project started by Museum Rotterdam in the Netherlands and now is part of the Stichting Wijkcollectie (District collection Foundation) programmes\footnote{\url{https://museumrotterdam.nl/ontdek/categorie/echt-rotterdams-erfgoed}, retrieved April 30, 2022.}\footnote{\url{https://wijkcollectie.nl/ere/}, retrieved April 30, 2022.}. The ERE project started with a Volkswagen minivan (shown in Figure~\ref{fig_minivan}) that belonged to a Bulgarian named Kamen. Kamen was travelling back and forth between Bulgaria and The Netherlands with this minivan, transporting his fellow Bulgarians who had been travelling between the two countries. Museum Rotterdam considered this minivan to be an embodiment of the changing Europe and city of Rotterdam. However, the question was how such an item could be `collected', while the owners still needed it for traveling across Europe. Museum Rotterdam decided to number the minivan and register it in their collection, but not to take it out of its running to place it in their depot. After identifying an initial collection of 30 items, Museum Rotterdam called on Rotterdam residents for additional proposals, and was overwhelmed by the number of proposals received.

The Museum then decided to put out a call for Rotterdam residents to form a collection committee. The committee has, since 2017, determined collection criteria and met regularly for collection meetings. To date, one hundred exemplars of Rotterdam's living heritage have been included in this collection, such as the Mayor Aboutaleb, who has been Mayor of Rotterdam since 2009, the Pauluskerk, a church that offers overnight shelters and warm food for those in need, and the Summer Carnival Rotterdam, reflecting the multicultural character of the harbour city.

In both examples, the museums rely on the local communities' insights into lives in their cities,
and explicitly let the communities identify what heritage matters to them. This raises attention to aspects that probably would not have been covered from the museum practitioners' perspectives alone.

\section{Reflecting on AI practice}\label{sec:ai_reflection}
Having described multiple key principles and actions observed in the museum community to promote social inclusion, we now reflect on ways in which these also can be useful for AI applications. Here, we especially consider AI applications that involve supervised machine learning procedures on data that relies on human interpretation, and where intended applications will involve curation steps.

\subsection{Neutrality Revisited}
As surveyed by Birhane et al., in many published research works in machine learning, generalisation and efficiency can be recognised as core important values defining achievement~\cite{values_in_ML}. Denton et al.~\cite{genealogy_datasets} point out the common (often implicit) stance that more data examples will lead to better learning outcomes. At the same time, dataset construction historically has been less valued and incentivised in the community, as also emphasised by Sambasivan et al.~\cite{data_cascades}. Thus, available large-scale datasets have commonly been reused, becoming `naturalised' and appearing value-neutral in the process~\cite{genealogy_datasets}.

The stance that datasets---and the technologies built on them---are not actually value-neutral, has only recently been gaining traction and debate in the machine learning community~\cite{ScheuermanEtAl2021datasetpolitics}. However, this notion has already been recognised as early as in 1996, in Forsythe's paper outlining insights from three years of anthropological observations in an AI development project for migraine sufferers~\cite{Forsythe1996}. The work illustrates the hidden presumptions in developing an AI tool, arguing that different decisions are made with different intentionality. Designers might think they cannot build their own cultural backgrounds and values into the system they are developing, but these explicit or tacit intentions often carry hidden presumptions. These tools are not only `located' in specific design settings, but also in specific practice settings. The people who deal with data will consciously or unconsciously be influenced by their own background and those of others. By claiming to be neutral or striving to be neutral, the chances of addressing the influences and biases would actually be lost. As argued by Scheuerman et al.~\cite{ScheuermanEtAl2021datasetpolitics}, such an awareness in documentation practices is key to changing the culture of dataset development among the practitioners.

Thus, bias is not necessarily problematic per se, and always will be present to some degree. Both professional archives~\cite{JoGebru2020archives} and museums will explicitly not seek to collect unbiased samples of the broad universe, but will set an explicit collection policy, purposefully making choices according to this policy and the institute's overall mission statement. As already argued by Jo and Gebru~\cite{JoGebru2020archives}, more transparency on the policies and criteria behind machine learning datasets can be achieved through more explicit policy documentation, and existing proposals on datasheets for datasets~\cite{data_sheets} and model cards for model reporting~\cite{model_cards} can naturally complement this. It will be beneficial to explicitly write these with a mindset of cultural humility, accepting up front that the currently presented world view is not the one and only, canonical world view, and perspectives may need adjustment over time.

It should be noted that many `canonical', heavily-employed, large-scale machine learning datasets were explicitly not constituted with a mindset of cultural humility. Multiple datasets in computer vision have been heavily criticised for including problematic and offensive imagery~\cite{pyrrhic_win} and labels~\cite{excavating_ai}, and displaying an Amerocentric and Eurocentric representation bias~\cite{geodiversity}.

The follow-up to these criticisms had various forms. Where the work of Shankar et al.\ on geodiversity issues both sparked more inclusive benchmarking efforts~\cite{inclusive_images} and well-publicised commentaries drawing attention to culturally specific AI biases~\cite{ai_sexist_nature}, Crawford and Paglen pointing out offensive class labels in ImageNet~\cite{excavating_ai} led to those categories being removed from the ImageNet public dataset. Even more radically, Birhane and Prabhu's audit~\cite{pyrrhic_win} of the Tiny Images dataset~\cite{tiny_images} led to this dataset to be fully and permanently taken offline.

As discussed in Section~\ref{neutralitymuseum}, it is not uncommon in museums that language and representation of collections items used to be accepted at collection time, but today would be considered derogatory and offensive. In reaction, this language has been revised and updated; however, this does not necessarily have to mean the previous language is fully erased. For example, the Words Matter document explicitly addresses these terms, but with clear contextualisation of how these terms today are problematic, and a clear underlying message that these terms should not be used anymore. Contrasting this to the removal actions undertaken on problematic machine learning categories and datasets, while full permanent removal prevents future harmful automatic inferences, it also fully cuts off the connection and context to earlier, existing models that were trained on these datasets.

\subsection{Situational Interpretation}
The previous discussion raised that over time, descriptive language may need revision, updating, and re-tailoring to the presently relevant audiences. In the discussion in Section~\ref{contextmuseum}, we raised that for museums, these audiences can be highly diverse, and in need of different languages and perspective highlights. These notions lead to multiple reflections on AI practice.

First of all, in museums, the same collection item may be described in different ways, or purposefully draw attention to different aspects of the item. When seeing a picture of a carriage driver with a top hat in Central Park, some people will focus on characteristics of the carriage, some will reminiscence about being a carriage driver, and some may remember a similar carriage trip elsewhere; these memories may be visual and factual, or abstract and affective. They are very differently-natured responses, that jointly constitute the experience to the item, while the physical representation of the item itself remains the same. In an AI context, the item would have the exact same encoded input representation, but a highly diverse vocabulary of associated output labels. Would these be considered in a single, general learning or tagging task, the same problems as encountered with ImageNet would occur~\cite{excavating_ai}, where not all terms would have a visual observable link, and some associations may not be desirable to learn.

Furthermore, considering the Isabella painting depicted in Figure~\ref{fig_isabella} and discussed in Section~\ref{neutralitymuseum}, without context, one may describe the painting as a depiction of a seated young girl with a fan in fine clothing. In contrast, a professional may describe it as an oil on canvas painting. However, knowing the context of the discussion this painting sparked, it may now be described as the art object that triggered the Rijksmuseum's terminology revisions, and used to be problematically titled. Again, this further strengthens the perspective that single, universal, `neutral' descriptions and representations of the item should not be strived for. Parallel vocabularies, labels and descriptions may validly exist for a single item, referring to different sources and broader background contexts. Encoding these as multiple tags of a single vocabulary, which often would happen in supervised machine learning contexts, would both be too simple, and taxonomically problematic.

In terms of encoding parallel, polyvocal interpretations, richer labels and annotations can be achieved by employing Web technology. The Semantic Web naturally gives opportunities for richer and parallel linking of differing viewpoints, as e.g.\ outlined by Van Erp and De Boer~\cite{cultural_ai_polyvocal}. Next to this, for digital music content, Solid Pods were proposed by Weigl et al.\ as a means to link various content types (e.g.\ textual annotations, scores, and recordings) at various levels of authority and intended visibility (e.g.\ professional commentary vs.\ private notes), facilitating provenance information, while explicitly keeping governance in the hands of the owner~\cite{read_write_libraries}.

Beyond technological questions of encoding and representation, more conceptually, questions can be asked about the nature and scope of personalisation. In search engines and recommender systems, personalisation typically happens in the back-end of a system as part of the filtering process, leading to personalised curated samplings of a collection. Different user profiles may be offered different selections of items to consume, that may also be differently ranked. 

An interesting notion in the museum practice is that the filtering process was performed by a human curator, and the filtered sample of a collection for a given exhibition will be fixed: everyone going to the same exhibition will in principle be offered the same collection items to see. Any possible personalised differentiation will be in the presentation and interpretation; the parallel in the digital world would be in the front-end user experience.

With regard to this, the discussion in Section~\ref{contextmuseum} very strongly emphasised purposeful communication practice, that would link elements of curation and education. In digital information access, tailored communication strategies focusing on item presentation are still relatively rare. Furthermore, in machine learning pipelines, deliberate connections across different stages in the process---from collection through curation to presentation---are hardly seen. Typically, datasets would be established in a general-purpose mindset: model learning may be tailored to a given learning task, but the user-facing presentation of the outcomes of this task are considered to be outside of the scope of the machine learning procedure. The equivalent to the audience-tailored museum exhibit experience, where end-user-specific front-end requirements would affect criteria on data annotation and curation, has to the best of our knowledge not been implemented as a process yet.

Especially in current works emphasising the importance of quality assurance on data, calls for a more holistic take are emerging and increasing, explicitly connecting practice downstream to dataset creation processes upstream. This e.g.\ is the case for the proposal of Hutchinson et al.\ to more explicitly include software development practices in dataset development~\cite{data_software}, but also for Sambasivan et al.'s work on data cascades, showing that uninformed choices made during dataset development will cause major problems in downstream applications~\cite{data_cascades}. In addition, it increasingly is acknowledged and recognised that more holistic connections will need to be made between machine learning prediction in AI and user experience (UX) considerations, e.g.\ in the call to action put forward by Cramer and Kim~\cite{tensions_ux_ai}.

\subsection{Community participation}
In striving for social inclusion, museums have emphasised a pro-active focus on communities whose voice has not yet been heard and uplifted. Translating this to AI practice, care should be taken in doing this. Often, minoritised communities were already implicitly involved in system development, but as low-paid laborers, conducting data labeling tasks during the dataset creation phase~\cite{genealogy_datasets}. This labeling procedure would normally not solicit their personal perspectives, but rather ask for standardised descriptions. Indeed, when participatory machine learning was coined as a way to more inclusively involve communities, warnings were put out by Sloane et al.\ that this should not lead to further exploitation and participation-washing, where community work does not benefit the community, but rather serves a party with power over this community~\cite{participation_not_design_fix}.

Looking at how the museum world implemented community participation, all of the examples discussed in Section~\ref{communitymuseum} clearly indicate agency of the community itself. In the ecomuseum setting, the items of relevance to the community explicitly remain in the context and under the governance of the community. When inviting community members to perform community creation, it also is clear the suggestions and insights will be respected, even if they may differ from what an `authoritative' curator would do.

Looking at participatory methods in the digital world, participation through a citizen science perspective may be a better form than through a crowdsourcing perspective. Where in the latter case, participant knowledge is solicited by a requesting party, to benefit that party's intentions, in the earlier case, the participant is an actual, fully acknowledged participant, who will more comprehensively join the full process leading to relevant insights. As e.g.\ documented in Hsu et al.~\cite{local_communities_citizen_science}, the community citizen science approach can effectively include communities, engaging on concerns that demonstrably are relevant to them.

Finally, the community participation practices in the museum world again illustrate the importance of communication and education. Earlier, in Section~\ref{neutralitymuseum}, it was emphasised that cultural humility may not naturally be ingrained in an institute, and minoritised communities will need to assist in educating the institutes about their blind spots. The other way around, coming back to Sambasivan et al.~\cite{data_cascades}, data cascades can be avoided when data literacy is improved, which presently still gets little attention and investment.

\section{Actionable organisational change: libraries in digital transition}

Having built forth on the Lessons from Archives paper~\cite{JoGebru2020archives}, and now discussing insights from the museum sector, we have been broadening perspectives on social inclusion and AI from a wider range of the so-called GLAM sector (galleries, libraries, archives and museums). While differently-natured in their provided services, all these four institution types are responsible for the collection, preservation, presentation and dissemination of cultural heritage.

To further illustrate the relevance of GLAM sector experiences for current social inclusion challenges in AI, as a final illustration, we wish to draw attention to the library sector. In terms of its practice and servicing, this sector may sit between the museum community and AI communities. Currently, it actively goes through a digital transition in which former core public values (e.g.\ `neutrality') need revision and debate. In showing how this currently is tackled in the public library community in The Netherlands, we again have an inspirational example of how social inclusion in collection and curation contexts is taken up as an organisational challenge, driven by explicitly set policies.

The public library has historically been a source for knowledge and information, purported to be offered in neutral ways. However, local public libraries never would have the shelf space to store all the books in the world. Therefore, collection policies have always actively been applied. With digitalisation boosting online information access, today, the public library may not be sought out as much anymore for knowledge. However, it still very explicitly is a social meeting space, where people from different backgrounds and perspectives may naturally encounter one another, where this would not be trivial anymore in other aspects of daily life. Against the backdrop of these developments, a dedicated Dutch law on public library facilities was passed in 2015~\cite{wsob}, quoting independence, reliability, accessibility, pluriformity and authenticity as core public values to uphold, and listing five core library tasks:

\begin{enumerate}
    \item to facilitate knowledge and information;
    \item to offer possibilities for development and education;
    \item to promote reading and introduce literature;
    \item to organise encounters and debates;
    \item to acquaint the audience with arts and culture.
\end{enumerate}

Looking at this, it is striking that `neutrality' indeed is not mentioned, but `pluriformity' is. Even more strongly, strategic library innovation agendas especially emphasise the pluriformity aspect~\cite{kb_kennisagenda}, noticing that the public library is one of the few remaining trusted places where confrontations with a different perspective than one's own may be tolerated.

Interestingly, the present eight ethical principles for using AI~\cite{van_wessel_jan_willem_2020_3865344}, authored by the National Library of the Netherlands, still call for neutrality as a core principle, explaining this as ``not steering users into directions that they may not choose outside of AI context''. At the same time, the libraries voice explicit interest in digital recommender systems, and a major concern voiced in strategic documents is that the application of AI in filtering processes would narrow down a user's perspective in undesirable ways~\cite{kb_kennisagenda}. In other words, the neutrality core principle needs revision, and responsible AI recommenders in the public library context maybe even \emph{should} actively steer users beyond the perspectives they would choose without the recommender.

Beyond the possible role of AI technology in this challenge, similarly to museums, libraries do not only seek to stimulate more inclusive practice as part of a filtering or book-lending procedure. Considering the five core tasks of a public library, organised social activities and cultural programming explicitly are included as part of these tasks, and actively employed to further engage relevant audiences. Here, the parallels with the community-oriented activities of museums can easily be seen.

In line with their task to offer possibilities for development and education, Dutch public libraries have today been designated as the place where citizens can learn important skills necessary for daily life. This includes courses on reading comprehension, but also increasingly focuses on digitalisation. Beyond introductory computer courses, this e.g.\ includes assistance with making use of digital public services, such as making vaccine appointments in the COVID-19 pandemic. Furthermore, the public libraries also will be conducting outreach projects to raise citizen awareness of AI. Increasing this awareness is vital for allowing more inclusive and informed public debates on AI technologies, their impact on society, and the true needs and wishes of users. In the United States of America, for similar reasons, the `We are AI' course was developed and implemented as a collaboration between the Center for Responsible AI at New York University’s Tandon School of Engineering, the Peer 2 Peer University, and the Queens Public Library, in the form of a learning circle at the public library\footnote{\url{https://dataresponsibly.github.io/we-are-ai/}, retrieved April 25, 2022.}.

\section{Conclusion}
In this article, we reviewed how social inclusion practices have been integrated in the practice of museums, particularly focusing on the perception of `neutrality', the need for situational interpretation, and the role of community participation. We highlighted how mindset changes in the museum world may be inspirational in tackling social inclusion challenges in AI technology.

We wish to emphasise that the museum sector did not yet `solve' matters of inequality in representation and ownership. With the ICOM museum definition discussion still ongoing, social inclusion has not formally been institutionalised yet. Furthermore, some societal public museum statements have been criticised to be lip service. For example, during the global Black Lives Matter movement in 2020---in the middle of the COVID-19 pandemic and anger surrounding the death of George Floyd at the hand of the Minneapolis police---major museums such as the British Museum issued statements to express solidarity with the movement. These gestures were, however, met with criticisms that museums need to do better to turn words into deeds by re-examining how Black communities have been portrayed. In the case of the British Museum, for example, the Museum came under fire again for controversial colonial-period objects within its collection, such as the Benin Bronzes~\cite{artnetBM2021,artnewsBM2021}.

Criticisms on museums following the Black Lives Matter movement show that the minority and underrepresented communities are not yet sufficiently satisfied, and that the museum sector is still a long way from resolving issues of social inclusion and cultural diversity in its practices. Still, at the same time, many relevant discussions are being held, many policies have noticeably been adjusted, and many tangible actions can be noticed. In our discussion of these, it should be noted that conscious attitude changes are addressed at different stages of the curation and presentation process, with a strong focus on tailored presentation and its surrounding communicative intent. Furthermore, the examples from the museum world also explicitly indicate how social inclusion requires strong organisational commitment, and how different departments within an organisation should actively be engaged in striving to improve it, while sustaining a mindset of cultural humility.

In the AI community, the rising awareness of socioculturally specific perspectives, interpretation dynamics, bias and power has so far largely been concentrated at the side of dataset creation and documentation~\cite{data_sheets, data_cascades, genealogy_datasets, data_software, bias_power}. However, in our current practice, commonly used datasets may get repurposed for different downstream tasks and applications, where the parties working on these tasks and applications may not always have been the dataset builders. Furthermore, considering a final application, user-facing experience considerations will likely be handled by a very different team than the team working on data-driven components of the application. Thus, more holistic quality assurance mechanisms will be needed.

Even more strongly, any communicative efforts focusing on application adoption and acceptance may explicitly happen beyond technical application development cycles. In our discussions of museum practice, we have highlighted that there are good reasons to actively make these different departments connect, also in the light of alignment with strategic mission statements. We illustrated how the public libraries community in The Netherlands is presently going through a similar transformation, involving a more profound organisational mission reorientation, that manifests in a broad spectrum of associated activities.

Having positioned our current discussion next to Jo and Gebru's Lessons from Archives~\cite{JoGebru2020archives}, we have further strengthened the argument that the AI community can learn from organisational best practices of GLAM institutes. These parties build upon long histories of experience with questions of curation, while at the same time being recognised as trusted, and in many cases public, spaces. Many of these organisations currently are going through digital transformations. While this is happening, they can provide excellent environments for piloting more holistic takes on socially inclusive AI technologies in practice.

\begin{acks}
This work is part of the `Recommender requirements for pluriform perspectives in local public libraries' project (NWA.1228.192.235) of the National Science Agenda (NWA) research programme, which is financed by the Dutch Research Council (NWO). The National Library of The Netherlands (Koninklijke Bibliotheek, KB) is a partner in this project. This partner role is that of a relevant domain stakeholder, and is neither financially compensated by the project grant, nor financially contributing to the project grant. In addition, Dr Liem's research time is supported through a personal Veni grant (016.Veni.192.131) of the NWO Talent Programme. The NWO and the KB had no role in the design and conduct of the research presented in this paper; access and collection of data; analysis and interpretation of data; preparation, review, or approval of the manuscript; or the decision to submit the manuscript for publication. The authors declare no other financial interests.
\end{acks}

\bibliographystyle{ACM-Reference-Format}
\bibliography{acmart.bib,cynthias_refs.bib}

\appendix

\end{document}